\pdfoutput=1

\documentclass[11pt]{article}

\usepackage[preprint]{acl}

\usepackage{times}
\usepackage{latexsym}

\usepackage[T1]{fontenc}

\usepackage[utf8]{inputenc}
\usepackage[ukrainian, arabic, english, bidi=default]{babel}

\usepackage{microtype}

\usepackage{inconsolata}

\usepackage{graphicx}

\usepackage{url}

 \usepackage{todonotes}
\usepackage{makecell} 
\usepackage{amsmath}

\usepackage{adjustbox}

\title{From English-Centric to Effective Bilingual: LLMs with Custom Tokenizers for Underrepresented Languages}

\author{
 \textbf{Artur Kiulian\textsuperscript{1}},
 \textbf{Anton Polishko\textsuperscript{1}},
 \textbf{Mykola Khandoga\textsuperscript{1}},
 \textbf{Yevhen Kostiuk\textsuperscript{1, 2}},
\\
 \textbf{Guillermo Gabrielli\textsuperscript{1}},
 \textbf{Łukasz Gągała\textsuperscript{1, 3}},
 \textbf{Fadi Zaraket\textsuperscript{6}},
 \textbf{Qusai Abu Obaida\textsuperscript{5}},
\\
 \textbf{Hrishikesh Garud\textsuperscript{4}},
 \textbf{Wendy Wing Yee Mak\textsuperscript{7}},
 \textbf{Dmytro Chaplynskyi\textsuperscript{}},
 \textbf{Selma Belhadj Amor\textsuperscript{7}},
\\
 \textbf{Grigol Peradze\textsuperscript{7}},
\\
\\
 \textsuperscript{1}OpenBabylon,
 \textsuperscript{2}ARG-Tech, University of Dundee, UK,
 \textsuperscript{3}Georg-August Universität Göttingen,\\
 \textsuperscript{4}Google,
 \textsuperscript{5}Arab Center for Research and Policy Studies,
 \textsuperscript{6}Doha Institute for Graduate Studies
 \textsuperscript{7}PolyAgent
\\
}

\begin{document}
\maketitle
\begin{abstract}
In this paper, we propose a model-agnostic cost-effective approach to developing  bilingual base large language models (LLMs) to support English and any target language. The method includes vocabulary expansion, initialization of new embeddings, model training and evaluation. We performed our experiments with three languages, each using a non-Latin script—Ukrainian, Arabic, and Georgian.

Our approach demonstrates improved language performance while reducing computational costs. It mitigates the disproportionate penalization of underrepresented languages, promoting fairness and minimizing adverse phenomena such as code-switching and broken grammar. Additionally, we introduce new metrics to evaluate language quality, revealing that vocabulary size significantly impacts the quality of generated text.
\end{abstract}

\section{Introduction}
The discovery of the Transformer architecture~\cite{attention} has opened doors for creating large language models (LLMs) with billions of parameters, trained on datasets of trillions of tokens. One of the notable features of the LLMs is cross-lingual language understanding (XLU), which allows models to possess multilingual capabilities. However, the XLU ability is restricted by the so-called \textit{curse of multilinguality}, which refers the difficulties and constraints encountered in creating multilingual LLMs. Studies showed that a substantial drop in performance occurs as the number of languages increases, due to the model's limited capacity to adequately capture and represent the nuances of each language~\cite{curse1}.
The efforts to examine and address the problem have highlighted two key factors: \textbf{the composition of the dataset} and \textbf{vocabulary composition}~\cite{curse2,curse3}. Some studies~\cite{curse4} suggest that the natural limitations on the model capacity, vocabulary and training dataset sizes along with differences in language structures do not allow the creation of the ultimate multilingual model to perform equally in many languages, favoring the creation of custom models targeted at specific languages instead.


The most obvious yet often overlooked consequence of low language representation in a model's vocabulary is a much higher cost of language processing. A sentence in Ukrainian requires about 3 times more tokens for the GPT-4 model~\cite{openai2024gpt4technicalreport} than the same sentence in English due to higher \textit{tokenization fertility} (see  Section~\ref{subsec:fertility}). Three times higher fertility means three times smaller context window, three times higher memory usage, and nine times higher computation cost due to attention's quadratic dependence on the sequence length.
On the other hand, high computational costs are not the only ramifications of a poor vocabulary.
Recent studies~\cite{howgood} indicate that representation in an LLM vocabulary of a specific language directly relates to the performance of the model in that language~\cite{tokenizer_bad}. In particular, it may be a reason for the generation of non-existing words, code-switching~\cite{codeswitch1,codeswitch2}, and broken grammar. Languages that use a non-Latin alphabet are particularly affected by poor vocabulary representation since they cannot rely even on the overlapping tokens with better represented languages.

An insufficient training dataset affects the performance of LLMs as much as it does any other deep learning model. The model might generate a response in the wrong language, probably the one it is most familiar with, such as English~\cite{langconfusion}. In this work, exposing the model to additional data in the target language via continual pre-training helped mitigate these effects.


In this paper, we present a model-agnostic resource-effective method to create a base bilingual LLM that supports English and another language. By addressing the above-mentioned issues of dataset and vocabulary composition, we make sure to improve its language capabilities along with boosting its computational efficiency. We illustrate our method in three languages with non-Latin alphabets: Ukrainian, Georgian, and Arabic.

The contributions of our work are as follows:
\begin{itemize}
    \item We propose a vocabulary extension procedure that preserves the model's accumulated knowledge of English and extends the target language comprehension.
    The method is verified with Gemma 2 and Mistral models (see Section~\ref{sec:vocab}).


     \item We trained two separate bilingual LLMs (English-Ukrainian and English-Arabic) on language-specific datasets using the Mistral~\cite{mistral} 7B model. The models were continually pre-trained for the next token prediction task on the parallel corpora for English and corresponding language. Our experiments showed that the proposed tokenization method reduces \textbf{computational complexity} and \textbf{inference time} for Ukrainian and Arabic respectively, while also improving model performance for code-switching and grammar correctness tasks. Additionally, we have conducted experiments to test the adoption of extended Georgian vocabulary for the English-Georgian model.

    \item We introduced new metrics for measuring code-switching and non-existing words ratio for Ukrainian and Arabic. The code-switching metric leverages the unique features of each language to detect instances of code-switching, following the rules of the respective languages.
\end{itemize}


\section{Related Work}

The shortcomings of existing multilingual LLMs have motivated numerous scholars and practitioners to address the insufficient performance of underrepresented languages.

Perhaps the most fundamental approach is to design and train a model from scratch, as demonstrated by EuroLLM~\cite{eurollm}. While this method offers maximal flexibility, it is highly demanding in terms of effort and computational resources.

More commonly, available open-source LLMs are used as a starting point, leveraging transfer learning and building on available weights~\cite{language_design}. This can still involve significant architectural changes compared to other methods, as seen in the SOLAR model~\cite{upstage}. Despite utilizing transfer learning, such approaches often require pre-training on vast datasets, sometimes reaching trillions of tokens.

A number of publications~\cite{chinese, tamil, vietnamese, vimistralx} suggest a more lightweight approach, where the model’s vocabulary is extended by 10,000–20,000 tokens, entailing the extension of the embedding layer and the language modeling head, while leaving the rest of the architecture unchanged. This method reduces the required training dataset to hundreds, or even tens, of billions of tokens, while still delivering notable improvements in the model's language abilities and computational efficiency.

Finally, instruction fine-tuning~\cite{italian, amharic, odia} offers a highly resource-efficient alternative by skipping the base model composition step. While this approach can yield some improvements, it does not enhance the model’s factual knowledge or address tokenization issues.

Our approach, in contrast, maintains the overall vocabulary size and keeps the model architecture intact. To create a bilingual model, we extend the vocabulary of the target language at the expense of other languages in the model, except English. This allows us to reduce the pre-training dataset to as little as 2 billion tokens while still improving the model's factual knowledge, enhancing the dataset, and achieving visible improvements in target language generation.





\section{Methodology}

Our proposed pipeline for training of bilingual LLMs supporting English and a target language $\mathcal{L}$ consists of the following steps:

\begin{enumerate}

    \item \textbf{Vocabulary Extension}. The aim of this step is to create a new bilingual tokenizer $T$ that retains the exact tokenization for English as in the original model, while incorporating an extended vocabulary for the target language $\mathcal{L}$, thus reducing fertility.


    \item \textbf{Embeddings Initialization}. Initialize new embedding vectors for the newly added $\mathcal{L}$-specific tokens. 
    
    \item \textbf{Continual model pre-training}. In order to allow the model to adopt the new tokens and use them during the text generation we have continually pre-trained the model with new extended vocabulary.


\end{enumerate}

Each step will be explained in more detail in the following subsections.

\subsection{Vocabulary Extension Methodology}
\label{sec:vocab}
In this paper, we experimented with Mistral and Gemma 2 tokenizers, which have vocabulary sizes of 32,768 and 256,000 tokens respectively. Both models use SentencePiece tokenizers~\cite{sentencepiece}.


Our vocabulary extension technique can be described as follows. Consider the original tokenizer $T_o$ that includes multilingual tokens. We trained a new tokenizer $T_{\mathcal{L}}$ for the target language $\mathcal{L}$ using a language-specific dataset. Next, the two tokenizer models are combined in order to obtain a bilingual tokenizer $T_{En-\mathcal{L}}$ that will be used during the training of the bilingual LLM. This is achieved via the following steps:
\begin{enumerate}
    \item In order to keep the English tokenization intact we copy all the English tokens from the original tokenizer model $T_o$ into bilingual tokenizer $T_{En-\mathcal{L}}$ along with their scores and IDs. We assumed that all tokens that contain only ASCII characters belong to English. We have also kept all the byte fallback tokens, control tokens (e.g. ``[SEP]''), and service tokens (e.g. ``[UNK]'').
    \item Tokens that belong in both $T_o$ and $T_{\mathcal{L}}$ are assigned IDs from $T_o$ and scores from $T_{\mathcal{L}}$. This procedure ensures tokenization according to the rules of $T_{\mathcal{L}}$ and at the same time allows the LLM to recognize familiar tokens of the target language $\mathcal{L}$ and to use the existing embeddings.
    \item Lastly, the vocabulary of $T_{En-\mathcal{L}}$ is filled with new tokens from $T_{\mathcal{L}}$ ensuring that the vocabulary size matches the original tokenizer $T_o$.
\end{enumerate}
The resulting bilingual tokenizer $T_{En-\mathcal{L}}$ is identical to $T_o$ in the tokenization of the English language. On the other hand, in the target language, its fertility is improved thanks to the extended vocabulary (see Table~\ref{tab:tokens_fertility}).

\subsection{Embeddings Initialization}
\label{sec:embeddings}

Upon the vocabulary extension, the embedding vectors for the new tokens must be reinitialized. A proper embedding initialization can significantly improve the training convergence speed, while failing to do so might lead to a slower convergence or even non-convergence~\cite{glorot2010understanding}.
In our experiments, we have tried a number of embedding initialization techniques, such as random, mean~\cite{mean_init}, FOCUS~\cite{focus} and technique we called \textbf{NA}tural \textbf{CH}aracter \textbf{O}verlap \textbf{S}egmentation (NACHOS). We selected NACHOS because it has shown better convergence during training (see Appendix~\ref{sec:app_embedds}). NACHOS works as follows.
New tokens in $T_{En-\mathcal{L}}$ are expressed through the tokens that have already existed in the original tokenizer model $T_{o}$. Every longer token $t_{new}$ can be split into a $n$ of shorter tokens $t$: $t_{new} \rightarrow (t_1...t_n)$, with shorter tokens belonging to the overlapping vocabulary. We then initialize the embeddings of these new tokens by computing the mean of the shorter tokens embeddings (see Eq.~\ref{eq:nachos}):
\begin{equation}
    E(t_{new}) = \frac{1}{n}\sum_{1}^{n} E(t_n),
    \label{eq:nachos}
\end{equation}
where $E(t_{new})$ represents the embedding vector of the new token, $E(t_n)$ denotes the embeddings of the overlapping token $t_n$ into which the new token is segmented.

\subsection{Continual pre-training}
As a final step, the newly composed model with the extended vocabulary and initialized embeddings is trained on the bilingual parallel corpora. This allows the model to fully adopt the new tokens, which we have verified by checking the token IDs of the model output. This process of new token adoption is put under scrutiny and discussed in detail in Section~\ref{subsec:token_adoption}.







\section{Datasets}
\label{sec:datasets}
\paragraph{Vocabulary Extension Datasets}The monolingual language-specific tokenization models $T_{\mathcal{L}}$ have been trained on monolingual datasets. For the Ukrainian language we've trained on the publicly available UberText 2.0~\cite{ubertext2}, that contains 3.274B words and consists of 8.59M texts.

To train an Arabic tokenizer we have used a private dataset of non-fiction books of 430 million words based on~\cite{doha_dict}.
For Arabic, we integrated one more additional preprocessing step. As an Arabic word could correspond to several words in another language transmitting the same meaning, it is the best practice to perform light stemming to allow the models to pick the similarity of the semantics of the main parts of words~\cite{arabic-stemm}. For example, we consider \foreignlanguage{arabic}{الـ} (English translation: \textit{the}) as a separate token when it prefixes a word. We processed attached pronouns and gender specifiers in similar way.

For our experiments with Georgian we have used the Georgian section of the public OSCAR dataset~\cite{oscar}, which contains 171.9M words. This dataset has been used for both tokenizer training and continual pre-training of the English-Georgian Mistral model for token adoption experiments.

\paragraph{Continual Pre-training Datasets} For continual pre-training we created parallel datasets, consisting of both English and target language.

For Ukrainian and Arabic, we considered Wikipedia parallel dataset dump from June 20th 2024 archive dump\footnote{
HIDDEN FOR REVIEW
}.
For Ukrainian, the size of the datasets is approximately 2B tokens. The total number of articles was 2.1M (791,336 in Ukrainian and 1,327,709 in English). The total number of Ukrainian tokens was 1.02B and the total number of English tokens was 1.05B.
For Arabic, the size of the datasets is approximately 1.8B tokens. The total number of articles was 2.1B (1.2B in Arabic and 882,534 in English). The total number of Arabic tokens was 621.51M and the total number of English tokens was 1.1B.
For Georgian token adoption experiments, we trained a model on parallel corpora from the same dump. The dataset was much smaller due to a sparsity of resources in Georgian. It contained 107,123 and 169,602 articles in English and Georgian, respectively. The total number of tokens was approximately 395.2M (219.88M in English and 175.32M in Georgian).

The articles were shuffled to create the training dataset with equal representation of the target language (Arabic or Ukrainian) and English. To determine the amount of tokens, we used the Gemma 2 tokenizer.



To evaluate the results, we used {FLORES-200}~\cite{nllb2022}\footnote{\url{https://huggingface.co/datasets/facebook/flores}} dataset for corresponding languages. The dataset is a collection of parallel translation corpora for 200 distinct languages, including Ukrainian and Arabic. We selected 500 text samples per language from the ``devtest'' split of the dataset in Arabic and Ukrainian. Each text was separated into tokens by space, and only initial 3 tokens were kept as a model input. Finally, these inputs were provided to the model to generate a completion with a maximum generated sequence length of 128.
For Ukrainian inputs, we obtained 1,500 tokens and 1,098 unique tokens. For Arabic inputs, we obtained 1,500 tokens and 1,000 unique tokens.

\section{Experimental Setup}
We continually pre-trained bilingual models on the next token prediction task on the parallel corpora utilizing HuggingFace~\cite{wolf2019huggingface, Tunstall_The_Alignment_Handbook} instructions for 8x80Gb GPUs. To launch training, we used the SkyPilot framework~\cite{skypilot}. In order to isolate the effects of extended vocabulary and additional pre-training we have conducted the same pre-training for the vanilla models and then compared the performances. For hyper-parameter optimization we used grid search. The selected set of hyper-parameters can be found in our GitHub repository\footnote{
HIDDEN FOR REVIEW
}.

\section{Evaluation Metrics}

Since we work on the base completion model, we focused mainly on the metrics that reflect the text completion performance: tokenizer fertility, code switching score, non-existing words ratio, and manually evaluated grammar correctness score.

\subsection{Tokenizer Fertility}
\label{subsec:fertility}

Fertility is the most common metric for evaluating tokenizer performance~\cite{fertility1,fertility2}. This is an intrinsic metric of the tokenization model and is defined as the average number of tokens required to represent a word. For a tokenizer $T$ and a dataset $D$, fertility is calculated by dividing the total number of tokens in $T(D)$ by the total number of words in $D$.




\subsection{Non-Existing Words Ratio (NEWR)}

We used a following heuristic to detect non-existent words generated by LLMs. A word is considered non-existent if it is absent from a large language-specific corpus or vocabulary. For Ukrainian, we used the Ubertext fiction corpus~\cite{ubertext2} to create a set of 2.6M unique words, mostly Ukrainian. Each generated word is checked against this set, and if absent, it is marked as non-existent. The Non-Existing Words Ratio (NEWR) was calculated as the percentage of non-existent words in the output for each language-specific LLM output.

Arabic requires more processing, as it is a language with several dialects associated with it. While each Arabic-speaking region has its own dialect, it significantly intersects with the modern standard Arabic (MSA), which is used in legal, news and other domains. While in this work we focused on MSA, dialectal words are often present in MSA. Therefore, we used the corpora associated with the Doha historical dictionary of Arabic~\cite{doha_dict}\footnote{\url{https://dohadictionary.org/}} to cover traditional Arabic~\cite{ArTopicACRPSBooks}, Aya Dataset~\cite{aya-arabic} to cover MSA, and Lisan corpora~\cite{lisan} to cover accepted dialectal words, 3.9M words in total.


\subsection{Code Switching Word Ratio (CSWR)}
In linguistics, code switching is a phenomenon, when a speaker uses (or ``switches'' between) two or more different languages in a conversation. To detect code switching in LLM outputs, we introduced a novel metric: Code Switching Word Ratio (CSWR). Unlike previous token-based methods~\cite{langconfusion}, our approach uses language-specific rules to better identify code switching. The implementations are available in the GitHub repository\footnote{
HIDDEN FOR REVIEW
}.

CSWR is a ratio of words in the text that includes at least one foreign symbol (outside of the alphabet of the language, not a number or punctuation) and does not fit the rules of the correct code switching usage. The lower this ratio is - the better performance model showed from a code switching perspective.

The correct instances of code switching are detected depending on the language. A detailed explanation and a list of rules are provided in the Appendix~\ref{sec:code_switching_rulesf}.

\subsection{Grammar Correctness Score (GCS)}

To evaluate grammar correctness, the model generated text was evaluated by experts for the particular language on the following criteria: usage of incorrect words (e.g. wrong gender of the word, plural and single word form confusion, non-existing words, word merging, typos etc.), incorrect capitalization and punctuation and instances of incorrect code switching. If any of those flaws were encountered by the annotator the score of 0 was assigned to the text. If the text passes the check, it was assigned the score of 1. Finally, the \textbf{Grammar Correctness Score (GCS)} is calculated as an average of all assigned scores for the test completions.

For each language (Ukrainian and Arabic) we employed three native speakers annotators.

\section{Results}
\subsection{Tokenizer Intrinsic Performance}
\label{subsec:tokenizer_performance}
The comparison of the original model tokenizer with the customized bilingual tokenizers developed by us via the procedure described in Section~\ref{sec:vocab} can be found in Table~\ref{tab:tokens_fertility}.
Besides Mistral with its 32,768 tokens in the vocabulary we have also experimented with Gemma 2, which has a vocabulary 8 times larger. That has allowed us to substantially extend the target language vocabulary without changing the model architecture.
Naturally, in every case the extended vocabulary has improved the tokenization fertility in the target language, allowing the model to process the same amount of text at lower computational cost. The non-linear fertility improvement is expected due to the logarithmic character of its dependence on the vocabulary size~\cite{optimal_tokenizer_size}.
\paragraph{Ukrainian}
In the case of the Ukrainian language, it was challenging to estimate the exact number of the language-specific tokens in the original vocabulary due to possible confusions with other languages that use the Cyrillic alphabet. The number presented in the Table~\ref{tab:tokens_fertility} is a lower estimate.
Fertility has been measured with 13 million words from the Ukrainian section of the OSCAR dataset.
Notably in the case of Gemma 2 we have developed a tokenizer that ensures the same fertility for the English and Ukrainian languages, thus reaching parity between the two. Parallel fertility has been measured using the Paracrawl parallel English-Ukrainian dataset~\cite{paracrawl}.
\paragraph{Arabic}
For the Arabic language, fertility was measured using a stemmed dataset (see Section~\ref{sec:datasets}). Due to this, the numerical fertility results for Arabic differ from those of the other languages and can't be directly compared to them.
\paragraph{Georgian}
The original Mistral vocabulary did not cover 6 letters from the Georgian alphabet, which has forced the model to resort to byte fallback (see also Section~\ref{sec:tok_adopt}), which affected the original model's fertility in Georgian. Extending the vocabulary by 5,500 tokens has allowed to improve token usage by nearly three times.
Due to Georgian dataset size limitations we were not able to properly train and evaluate a Gemma-compatible tokenizer for the Georgian language.

\subsection{Token Adoption Process} \label{sec:tok_adopt}
\label{subsec:token_adoption}
In this subsection, we investigate the token composition of the Mistral model output during the continual pre-training that followed the vocabulary extension for Ukrainian (Mean initialization), Georgian (NACHOS initialization), and Arabic languages respectively. The output tokens have been split into 5 categories:
\begin{itemize}
    \item Existing: tokens of the target language that exist in the default Mistral vocabulary.
    \item New: tokens of the target language that were added to the vocabulary.
    \item English: tokens used to represent English.
    \item Byte-encoded: 256 byte fallback tokens used to encode characters absent in the vocabulary in UTF-8 format.
    \item Other: tokens that do not belong to any of the above-mentioned categories (e.g. tokens of other languages, punctuation, etc.).
\end{itemize}

\begin{figure*}[h!]
    \centering
    \includegraphics[width=\textwidth]{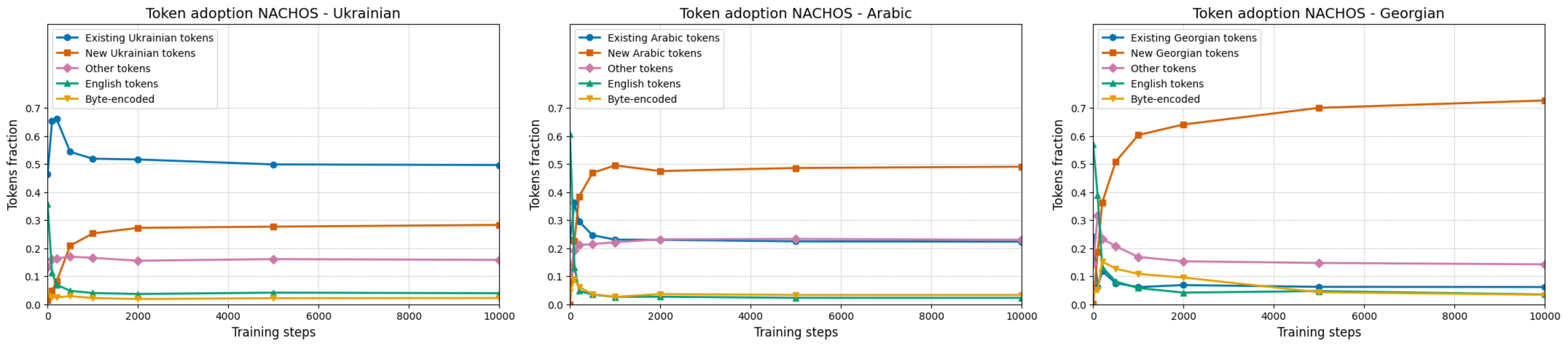}
    \caption{Tokens adoption by Mistral model.}
    \label{fig:mistral_token_adoption}
\end{figure*}

On Figure~\ref{fig:mistral_token_adoption}, Y axis of the plot corresponds to the relative fraction of the tokens in each category (all categories sums up to 1).
In general, we observed similar phenomena in all three languages.
Being prompted in a target language, the original Mistral model is likely to produce a response in English, most probably due to insufficient pre-training on the target language corpus. Once our pre-training starts, the model learns to produce responses in the target language and after a few hundred training steps it outputs little to no English tokens.

At first, the model favors the usage of familiar tokens that already existed in its vocabulary before the extension. Subsequent pre-training teaches the model to use the new tokens along with the familiar ones. After 2,000 training steps, the process stabilizes and becomes nearly static between 5,000 and 10,000 steps.

The same pattern holds in all three of the considered languages, though with some differences which we would like to discuss in more detail. We experimented with Ukrainian, Georgian, and Arabic.

\paragraph{Ukrainian}
Ukrainian is much better represented in Mistral model than Arabic and Georgian.
The original Mistral vocabulary contains 1,731 Cyrillic tokens, with about 1,600 of them suitable for the Ukrainian language representation. The original model occasionally replies in English if prompted in Ukrainian, producing about 35\% of English tokens in the output.
Upon the start of the pre-training the model learns to use Ukrainian tokens, though initially the model tends to use the existing Ukrainian tokens. After 200 training steps, this ratio increases to about 65\%. With more training, this number drops to 50\%, indicating that the model fully adopted new tokens. However, despite the new tokens make about 75\% of the extended Ukrainian vocabulary, the fraction of existing tokens remains dominant due to higher frequency of occurrence.

\paragraph{Arabic}
Qualitatively, the situation with the Arabic language is similar to that of the Ukrainian, but with two important differences. When prompted in Arabic, original Mistral is more likely to respond in English, with the fraction of produced English tokens reaching 60\%. In the original Mistral vocabulary there is 70 Arabic tokens, which is enough to avoid byte fallback, but is still a relatively small number. That is why the fraction of the new tokens overtakes as early as 200 training steps and remains dominant afterwards.

\paragraph{Georgian}
There are 29 Georgian tokens in the original Mistral vocabulary, which does not even cover the Georgian alphabet (35 letters). That forces the model to resort to byte fallback when generating text in Georgian more frequent than in Ukrainian or Arabic. The fraction of the byte encodings grows when the model learns to respond in Georgian and then drops along with the adoption of the new tokens, similarly to previously discussed languages. In case if Georgian, the token adaptation takes longer, as the model resorts to using the byte encodings for the text prediction while learning new tokens. Byte encodings are always encoded with a pair of tokens and that might explain a longer period of adopting the new Georgian tokens.

\subsection{Performance Metrics}

The results for the trained model of Grammar Correctness Score (GCS), Non-Existing Words Ratio (NEWR), and Code Switching Word Ratio (CSWR) are presented in Table~\ref{tab:perf_results}.

\begin{table}[]
    \centering
    \begin{tabular}{|c|c|c|c|}
    \hline
    \textbf{Model} & \textbf{GCS} $\uparrow$ & \textbf{NEWR}  $\downarrow$ & \textbf{CSWR} $\downarrow$ \\\hline\hline

    \textbf{Ukrainian} &  &  &   \\\hline
         Vanilla & 0.264 & 0.089                 & 0.515 \\\hline
         Tuned   & 0.388      & 0.032                 & 0.002\\\hline
         \textit{Ours}    & \textbf{0.503} & \textbf{0.030} & \textbf{0.001}\\\hline
         \hline

         \textbf{Arabic} &  &   &   \\\hline
         Vanilla         & 0.040                 & 0.863          & 0.450 \\\hline
         Tuned           & 0.238                 & 0.079          & 0.004 \\\hline
         \textit{Ours}   & \textbf{0.548}        & \textbf{0.050}  & \textbf{0.002}\\\hline



    \end{tabular}
    \caption{Results for trained model of Grammar Correctness Score (GCS), Non-Existing Words Ratio (NEWR), and Code Switching Word Ratio (CSWR). ``Vanilla'' refers to the original Mistral 7b model without additional training, ``Tuned'' refers to the continually pre-trained Mistral model on the same datasets, ``Our'' refers to Mistral continually pre-trained with extended vocabulary. $\uparrow$ indicates that bigger value is better. $\downarrow$ indicates that lower value is better. }
    \label{tab:perf_results}
\end{table}

The results showed that the model trained with our approach outperformed both Vanilla and Tuned models in terms of GCS in Ukrainian and Arabic. Notably, the vanilla model struggled with grammatical accuracy, achieving a score of 0.264 on Ukrainian compared to the our model's score of 0.503. Tuned English-Ukrainian model achieved GCS of 0.388. For Arabic, tuned model achieved 0.238 and 0.04 for the vanilla model, demonstrating lack of grammatical knowledge. Our model achieved GCS score of 0.548.

Our method demonstrated NEWR of 3\%, which is not significantly different from the score of the tuned model (3.2\%) for Ukrainian. The reason for such similarity could be in a better representation of Ukrainian tokens in Mistral (see Figure~\ref{fig:mistral_token_adoption}). Vanilla model showed 8.9\% of non-existing words in its generated texts. On the other hand, for Arabic our approach obtained NEWR of 5\%, when vanilla and tuned models obtained 86.3\% and 7.9\% respectively. The vanilla model's performance was really poor when it comes to generating existing modern Arabic words. The tuning improved the performance in more than 10 times, but our model outperformed it.

Finally, we achieved a score of 0.001 for CSWR for Ukrainian, which indicates a very little incorrect usage of foreign languages in the text. The second best score was obtained for tuned model (0.002). The vanilla model performed significantly worse: 0.515, indicating that more than half of generated words are used incorrectly in terms of code switching. For Arabic, the situation is similar. Our model obtained a score of 0.002, outperforming tuned model (0.004) and vanilla model (0.45).

\subsection{Preventing catastrophic forgetting in English}
After a series of experiments, we found that after just 1 epoch of training on the bilingual corpora, the models showed improvement in the target language but experienced a substantial drop in the English MMLU benchmark~\cite{mmlu, mmlu-ethics}. However, by lowering the learning rate from $1.5e-5$ to $2e-6$, training resulted in a much smaller loss in MMLU benchmark points. These important results demonstrate that, with the right training, the model can retain its English performance and remain bilingual, as shown in Table~\ref{tab:low_lr}.

\begin{table}[h!]
\centering
\small
\begin{tabular}{|c|c|c|c|c|}
\hline
 Mistral& \multicolumn{2}{c|}{Vanilla} & \multicolumn{2}{c|}{\textit{Ours}} \\ \hline
\textbf{} & \textbf{Tokens} & \textbf{Fertility} & \textbf{Tokens} & \textbf{Fertility} \\ \hline
\textbf{Ukrainian} & 1,077 & 3.35 & 5,552 & 2.55 \\ \hline
\textbf{Arabic}* & 70 & 3.3 & 3,618 & 1.68\\  \hline
\textbf{Georgian} & 29 & 7.61 & 5,531 & 2.68 \\ \hline \hline
 Gemma&  \multicolumn{2}{c|}{Vanilla} & \multicolumn{2}{c|}{\textit{Ours}} \\ \hline
\textbf{} & \textbf{Tokens} & \textbf{Fertility} & \textbf{Tokens} & \textbf{Fertility} \\ \hline
\textbf{Ukrainian} & 6,426  & 2.55 & 75,704 & 1.16 \\ \hline
\textbf{Arabic}* & 6,075 & 1.65 & 32,333 & 1.52 \\ \hline
\end{tabular}
    \caption{Tokenization metrics. *Stemmed tokenization for Arabic.}\label{tab:tokens_fertility}
    \label{tab:tokenizer_metrics}
\end{table}

\begin{table}[ht]
\centering
\small
\begin{tabular}{|l|c|c|c|c|}
\hline
\textbf{Model} & \textbf{GCS}$\uparrow$ & \textbf{NEWR} $\downarrow$ & \textbf{CSWR}$\downarrow$ & \textbf{MMLU}$\uparrow$ \\
\hline
Vanilla & 0.26 & 0.09 & 0.52 & 0.59 \\
\hline
Tuned & 0.39 & 0.03 & 2e-3 & 0.34 \\
\hline
Ours & 0.50 & 0.03 & 1e-3 & 0.25 \\
\hline
Tuned$\dagger$ & 0.31 & 0.03 & 2e-3 & 0.49 \\
\hline
Ours$\dagger$ & 0.42 & 0.03 & 9e-4 & 0.507 \\
\hline
\end{tabular}
\caption{Retention of the MMLU performance in the English-Ukrainian models trained with low learning rate (denoted with $\dagger$).}
\label{tab:low_lr}
\end{table}

\section{Discussion}

The obtained results highlight a subject that has been largely overlooked, particularly in the context of generative LLMs: the impact of vocabulary size and composition an on the quality of generated text.

Our experiments with the vanilla model pre-training demonstrated that the effects of training on additional data can be mitigated via the vocabulary extension. 
Additional pre-training on the target language corpus can noticeably increase text quality, particularly in addressing issues like code-switching and the generation of non-existent words. However, handling more complex linguistic features, such as grammar, requires vocabulary extension. Ukrainian and Arabic tokens are represented differently in the original model’s vocabulary, resulting in distinct yet complementary outcomes for the two languages. While for Ukrainian a substantial 29.6\% improvement was obtained with the extended vocabulary, the severely underrepresented Arabic achieves a much higher 90.5\% improvement. This effect was confirmed with another round of training at a lower learning rate for the English-Ukrainian models, which showed a 35\% improvement utilizing the model vocabulary extension. 

We propose the following explanation for this phenomenon: a poor vocabulary results in tokens that contain only one or a few characters, conveying very little specific semantic meaning. As a result, the model is forced to rely heavily on context during training and inference. This increases the noisiness of the data and prevents the model from learning nuanced meanings or effectively constructing complex grammatical structures.

Unfortunately, a static and limited vocabulary with fixed token-to-embedding mappings is an inherent limitation of the transformer architecture. This makes it challenging to create a transformer-based LLM that is equally proficient in multiple distinct languages.

For this reason, we advocate training bilingual models, which are both cost-effective and proficient in their target languages.

\section{Conclusions}

In this work, we introduced a model-agnostic, cost-effective method for developing bilingual base completion LLMs that support English and a target language, including low-resource or underrepresented languages. Our approach, centered on vocabulary extension and efficient embedding initialization, was validated by creating two bilingual LLMs: English-Ukrainian and English-Arabic. Moreover, we conducted experiments with Georgian tokenization and explored token adoption process during the training of a English-Georgian model. Georgian has a unique underreprsentation in the Mistral tokenizer.

We demonstrated that extending the vocabulary of a pre-trained model enhances its performance in target language while maintaining its English performance. Specifically, the grammar correctness results indicate that pre-training alone provides only limited improvement. The comparison between Ukrainian and Arabic further emphasizes the limitations of poor vocabulary for the underrepresented language. Expanding the tokenizer's vocabulary with target language tokens reduced tokenizer fertility, resulting in lower computational costs and improved processing efficiency. Finally, retaining the original English tokens in the custom tokenizer while adding new language-specific tokens lead to preservation of the model’s English performance on the MMLU benchmark, while also improving its performance in the target language from perspective of grammar, code switching and non-existant word ratios.

Our approach promotes a more equitable and inclusive NLP ecosystem, contributing to the revitalization of underrepresented languages. By lowering the barrier to developing more literate and grammatically capable models, we believe our work also paves the way for enhanced economic viability of using LLMs in non-English languages.

\section{Limitations}

In this work, we have focused on creating a minimal working example of a base bilingual model with an extended vocabulary in a cost-effective way. While our approach is model-agnostic, it has yet to be tested with models other than Mistral 7B. Gemma 2 is the most likely candidate, as we have already concluded tokenizer experiments. However, applying the method to other open-source models, such as Llama 3 or Qwen, would provide further validation for our approach.

Another important limitation is that the method was eventually tested only for English-Ukrainian and English-Arabic models. Due to the limited availability of Georgian corpora, we were unable to complete the experiment with the English-Georgian model. 

The retention of English language capabilities has only been tested with the English-Ukrainian model. We are currently in the process of testing it for the English-Arabic model. 

Further experiments with the vocabulary size and composition could help to find the optimal parameters along with their dependence on the available dataset size and individual language properties.

To fully evaluate the model across a variety of downstream tasks, such as machine translation, question answering, summarization, or text completion, instruction tuning will be required. This step, however, goes beyond the scope of our current work.

While we believe that the proposed metrics for assessing the language quality are an important step, they leave enough space for refinement. In particular, the code-switching metric for Ukrainian and Arabic might benefit from implementing additional rules.

\section*{Acknowledgements}
We would like to express our gratitude to the following organizations for their generous support, which made this work possible:
\begin{itemize}
\item \textbf{Observea} for providing access to a 16xTesla H100 cluster, which significantly enhanced our computational resources.
\item \textbf{NVIDIA} for providing DGX Workstation with 4xTesla V100 used for inference and evaluations.
\item  \textbf{HotAisle} for granting access to the 8xAMD MI300x node, enabling critical model training experiments.
\item \textbf{AWS} for offering cloud credits that supported the use of Tesla H200 instances for training.
\item \textbf{GCP (Google Cloud)} for providing credits used for model training and inference.
\item \textbf{TPU Research Cloud (Google Cloud)} for providing TPU VMs for our training pipeline experiments.
\item \textbf{A.I. Hero} for providing access to a 8xA100 instance for our original set of experiments.
\item \textbf{Doha Graduate Studies University} and the \textbf{Arab Center for Research and Policy Studies} for being key strategic collaborators, whose guidance and expertise greatly contributed to the development of this work.
\end{itemize}
All of the contributions above were instrumental in achieving the results presented in this paper, and we look forward to continued collaborations.

\bibliography{acl_latex}

\appendix

\section{Embedding Initialization Comparison} \label{sec:app_embedds}

\begin{figure}[!hb]
    \centering
    \includegraphics[width=\linewidth]{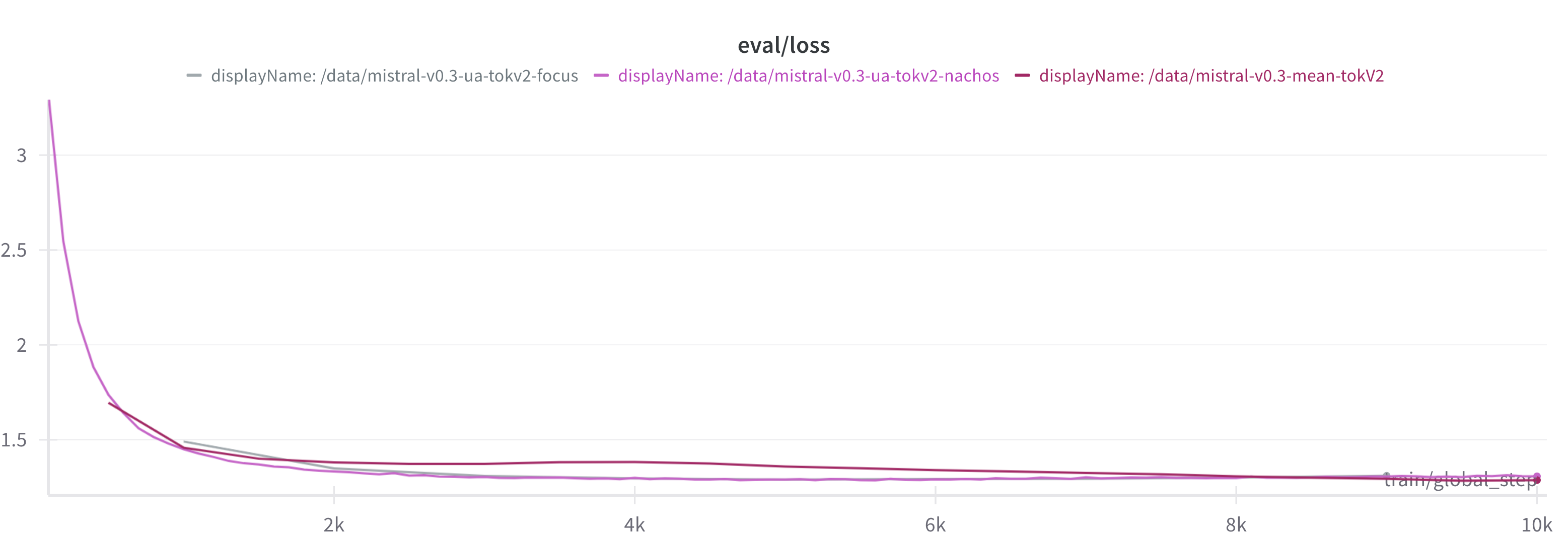}
    \caption{Ukrainian evaluation graph per training step. The name includes the embedding initialization technique: mean, residual, and NACHOS.}
    \label{fig:losses_app_ue}
\end{figure}

\begin{figure}[!hb]
    \centering
    \includegraphics[width=\linewidth]{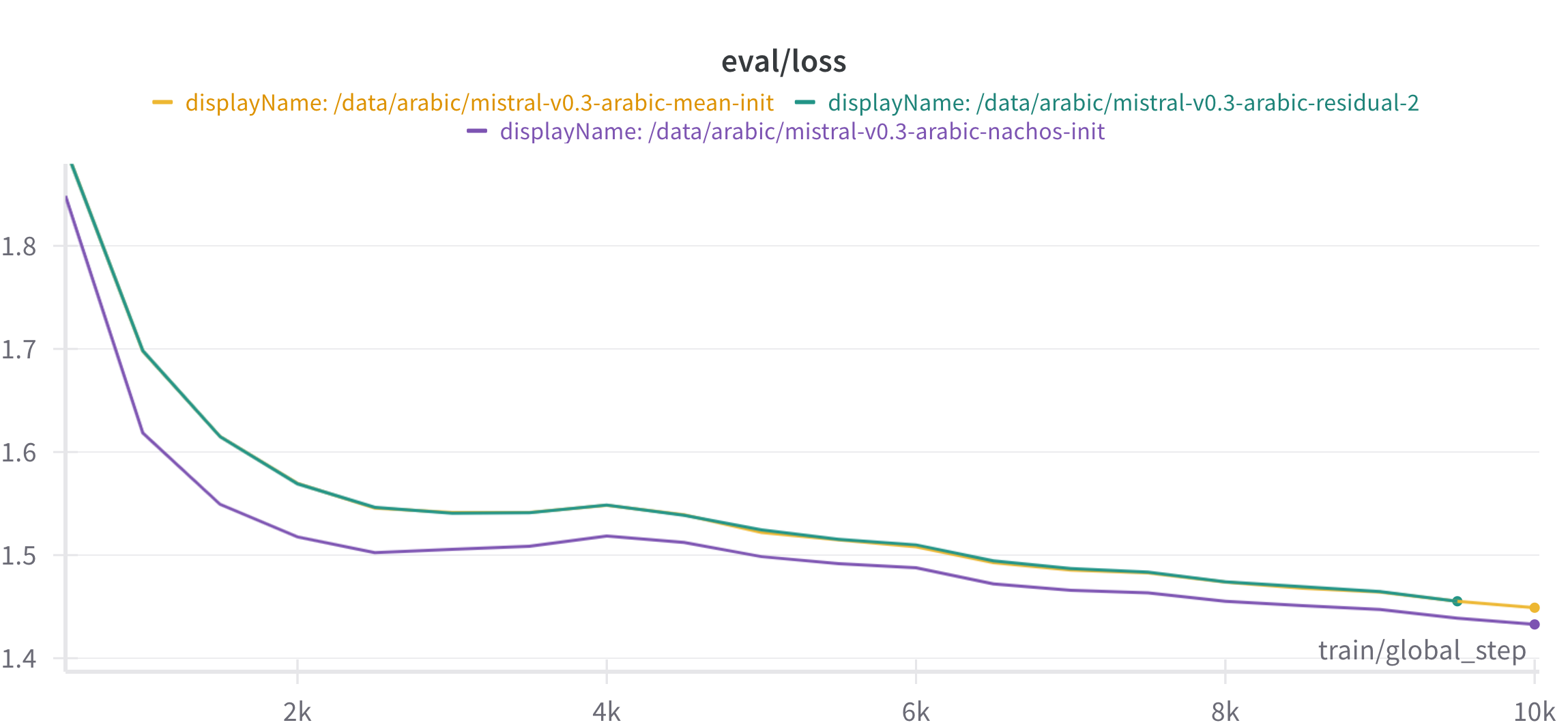}
    \caption{Arabic evaluation graph per training step. The name includes the embedding initialization technique: FOCUS, NACHOS, and mean.}
    \label{fig:losses_app_ae}
\end{figure}

In the Table~\ref{tab:embedd_app_comp} the metrics for the different languages and embedding initializations are presented. The graph of training and evaluation losses are presented on the Figure~\ref{fig:losses_app_ae} and~\ref{fig:losses_app_ue}.

\begin{table}[h!]
\centering
\begin{tabular}{|l|c|c|}
\hline
\textbf{Model} & \textbf{NEWR}$\uparrow$ & \textbf{CSWR}$\downarrow$ \\ \hline
Vanilla & 0.9118 & 0.5156 \\ \hline
Tuned & 0.9667 & 0.0006 \\ \hline
Mean & 0.9667 & 0.0009 \\ \hline
NACHOS & 0.9665 & 0.0009 \\ \hline
FOCUS & 0.9634 & 0.0011 \\ \hline
\end{tabular}

\caption{Comparison of Model Performance on NEWR and CSWR Metrics}]\label{tab:embedd_app_comp}
\end{table}
In our experiments, NACHOS demonstrated a better convergence compared to other methods, however the performance results for the final models were similar. As 
complete evaluation is computationally expensive and requires manual annotation, we decided to continue only with NACHOS approach. 

\section{Code Switching scoring rules} \label{sec:code_switching_rulesf}

To calculate the score for each language, the same initial preprocessing for the generated text was applied: the accents were replaced with regular corresponding letters and HTML formatting tags were removed.

\subsubsection{Ukrainian CSWR Rules}
In Ukrainian, the usage of code switching is allowed if it respects the following rules. All the mentions of the following entities are allowed in a foreign language:
\begin{itemize}
\item Proper names: names of the music bands, locations, restaurants, libraries, cities, titles, identification numbers etc. For example, \textbf{Pythagoras}, \textbf{California}, \textbf{MIT}, \textbf{Metallica}, \textbf{F-16} and so on.
\item Medical terms, additives and vitamins. For example, (vitamin) \textbf{B12}, (food additive) \textbf{E110} etc.
\item Roman integers and math symbols. For example, \MakeUppercase{\romannumeral 2}, \MakeUppercase{\romannumeral 10}, $\sum_{i=1}^N$, etc.
\item Quotes. If text is a direct quote, it can be used in Ukrainian without translation, marked with the special symbols.
\item URL links, hashtags, encoding names, mentions of the most common file formats and filenames. For example, \textbf{PDF}, \textbf{my\_cv.pdf}, \textbf{mydog.png}, \textbf{\url{https://www.wikipedia.org}}, \textbf{UTF-8}, \textbf{\#Euro2012} and so on.
\item Common Latin phrases. Some of the well-known Latin sayings and quotes can be used as is if they are widely known. For example, \textbf{Veni, vidi, vici}, \textbf{A priori} etc.
\end{itemize}

To accommodate these rules, our metric utilizes an ensemble of named entity recognition (NER) models as well as a rule-based approach to pick up the correct usage of foreign words or symbols. In particular, we have used XML-based Ukrainian NER model\footnote{\url{https://huggingface.co/EvanD/xlm-roberta-base-ukrainian-ner-ukrner}}, SpaCy~\cite{honnibal2020spacy} uk\_core\_news\_lg\footnote{\url{https://spacy.io/models/uk\#uk\_core\_news\_lg}} model, and Stanza~\cite{qi-etal-2020-stanza} Ukrainian model. All the URL links, Roman integers, math symbols, and text in quotation marks were extracted as separate named entities with the regular expressions. Finally, each sentence were checked if it contained any char in Ukrainian. If it did not and the whole sentence was not considered to be a named entity, the whole sentence and words in it were considered as incorrect.


To accommodate medical terms, additives and vitamins usage rule, we manually extracted a list of them from the US Food and Drug Administration\footnote{\url{https://www.fda.gov/food/food-additives-petitions/food-additive-status-list}}, as they can be used in Ukrainian language as well without translation. The total number of terms is 2,729.

To extract encoding names, file formats and filenames, and widely recognised Latin phrases, we manually retrieved them from Wikipedia. We obtained a list of 79 encoding names, 1,995 file formats, and 2,373 Latin phrases.

All of the resources are available on our GitHub repository\footnote{Hidden for review
}.

\subsubsection{Arabic CSWR Rules}

Arabic follows the following rules.

\begin{itemize}
\item Arabic does not have capital letters which renders named entity detection especially for proper names a specialized task.
\item In Arabic, both Indian or Arabic numerals can be used.
\item Some Arabic characters are non-connecting characters and are written separately from the next word, even if there is no space between them.  Arabic is written right to left, but Arabic words followed by non-Arabic words written in the other direction (sometimes  with no white space separation).
\end{itemize}

To address these issues, we utilized a different ensemble of NER models, specifically Flair~\cite{akbik2019flair} pre-trained Arabic NER model~\cite{phdthesis-arabic}\footnote{\url{https://huggingface.co/megantosh/flair-arabic-multi-ner}}, transformer-based Arabic NER models~\cite{lan2020gigabert, inoue-etal-2021-interplay}\footnote{\url{https://huggingface.co/ychenNLP/arabic-ner-ace}, \url{https://huggingface.co/CAMeL-Lab/bert-base-arabic-camelbert-mix-ner}}, and Stanza~\cite{qi-etal-2020-stanza} Arabic model.
Resources and algorithms to identify medical terms, additives, vitamins, hashtags, encoding names, URL links, file formats, roman integers and quotes are the same as we introduced in the Ukrainian Code Switching Metric.

\end{document}